\documentclass[conference, 9pt]{IEEEtran}

\usepackage{cite}
\usepackage{amsmath,amssymb,amsfonts}
\usepackage{algorithm}
\usepackage[noend]{algorithmic} 
\usepackage{algorithm}
\usepackage{graphicx}
\usepackage{graphics}
\usepackage{textcomp}
\usepackage{xcolor}
\usepackage{footmisc}
\usepackage{float}
\usepackage{booktabs}
\usepackage{url}

\begin{document}

\newcommand{\secspace}{\vspace*{0pt}}
\newcommand{\subsecspace}{\vspace*{0pt}}
\newcommand{\figspace}{\vspace*{-5.5pt}}
\newcommand{\figtitlespace}{\vspace*{-4pt}}

\pagestyle{plain}

\title{ELF: \underline{E}fficient \underline{L}ogic Synthesis by Pruning Redundancy\\in Re\underline{f}actoring}

\author{
    \IEEEauthorblockN{
        Dimitris Tsaras\textsuperscript{†,‡},
        Xing Li\textsuperscript{‡},
        Lei Chen\textsuperscript{‡,*},
        Zhiyao Xie\textsuperscript{†},
        Mingxuan Yuan\textsuperscript{‡}
    }
    \IEEEauthorblockA{\textsuperscript{†}The Hong Kong University of Science and Technology, Hong Kong}
    \IEEEauthorblockA{\textsuperscript{‡}Noah's Ark Lab, Huawei, Hong Kong}
}

\maketitle
\renewcommand*{\thefootnote}{\fnsymbol{footnote}}
\footnotetext[1]{Corresponding author: Lei Chen (\textit{lc.leichen@huawei.com})\vspace{4pt}\\
This work has been accepted for publication in DAC 2025. The final published version will be available via IEEE Xplore.}

\begin{abstract}
In electronic design automation, logic optimization operators play a crucial role in minimizing the gate count of logic circuits. However, their computation demands are high. Operators such as refactor conventionally form iterative cuts for each node, striving for a more compact representation - a task which often fails 98\% on average. Prior research has sought to mitigate computational cost through parallelization. In contrast, our approach leverages a classifier to prune unsuccessful cuts preemptively, thus eliminating unnecessary resynthesis operations. Experiments on the refactor operator using the EPFL benchmark suite and 10 large industrial designs demonstrate that this technique can speedup logic optimization by $3.9\times$ on average compared with the state-of-the-art ABC implementation.
\end{abstract}


\secspace
\section{Introduction}
\secspace

In logic synthesis, Boolean resynthesis is a technology-independent process that optimizes a logic network to reduce its nodes and levels~\cite{misch_2006, misch_2007, brayton_2007, amaru_2015_maj, soeken_2017, chu_2019, riener_2022}. It analyzes the Boolean expressions and iteratively applies local transformations (e.g., operators) to the logic gates and their interconnections. By leveraging these transformations, Boolean resynthesis optimizes digital circuits for lower power consumption, delay, and area. While Boolean resynthesis operators are essential for circuit optimization, they must be used with caution due to their computational expenses.

Since most resynthesis operators are heuristic and suboptimal, iterative runs are necessary to identify further potential improvements in the logic circuit \cite{misch_2006}. However, each successive iteration yields diminishing returns in terms of area improvement, while increasing the computational cost of logic synthesis. Moreover, the demand for a shorter turn-around time keeps increasing in the fast-paced electronics industry. By reducing the runtime of key optimization operators, designers can efficiently explore a wider range of design options, leading to better design solutions. Such an improvement of synthesis operators is design-agnostic, benefiting the synthesis process of almost all digital designs. 

In recent years, machine learning (ML) has been increasingly leveraged to enhance logic synthesis. Neto et al.~\cite{neto_2021} employ ML to reinvigorate logic optimization algorithms and explore more advanced designs. Other recent research works utilize ML techniques for sequence tuning to improve the PPA~\cite{yu_2018, yu_2020, grosnit_2022, li_2022, feng2022batch}. However, none of the ML-driven solutions improve the efficiency of the operators, which are the runtime bottleneck for realistic industrial designs with millions of nodes. For example, executing the refactor operator on a synthetic circuit with 23 million nodes in the EPFL benchmark suite \cite{amaru_2015} can take up to 1.5 hours. 
Despite some recent efforts in GPU acceleration~\cite{possani2018unlocking,lin2022novelrewrite}, there might be scenarios with hardware constraints that do not support such parallelism. Also, such GPU parallelism is potentially compatible with other improvements in operators. To the best of our knowledge, no prior work has addressed the redundancy in refactor or proposed a method to prune the unnecessary iterations. In this work, we propose ELF, a novel and general method, which prunes the redundancy in logic optimization operators and significantly accelerates the refactor operator. Our contributions are summarized as follows:
\begin{itemize}
    \item We present the redundancy problem in logic refactoring, and we provide a detailed examination of redundancy in both academic and industrial designs, which, on average, exhibit a failure to improve a node's cut by $98\%$.
    \item Then, we propose an efficient and accurate model to detect redundant operations during the refactoring process. Our model achieves an average accuracy of $87\%$ and recall of $93\%$ on academic circuits, and an average accuracy of $85\%$ and recall of $95\%$ on industrial designs.
    \item Finally, we propose ELF, an efficient logic refactor \cite{misch_2007} operator by integrating the redundancy detection model in the synthesis process. Compared with the state-of-the-art original ABC \cite{abc_2010} implementation, our method achieves an average speedup of $5.29\times$ with a minimal design quality loss of less than $0.27\%$ on academic designs. Similarly, on industrial designs, our method achieves a speedup of $2.80\times$ with a negligible design quality loss of less than $0.08\%$.
\end{itemize}

The remainder of this paper is structured as follows. Background and related work are reviewed in Section \ref{sec:background}. Section \ref{sec:method} provides an in-depth discussion of our proposed technique. Experimental results are presented in Section \ref{sec:exp results}, followed by concluding remarks and future directions in Section \ref{sec:conclusion}.

\secspace
\section{Background \& Related Work}\label{sec:background}
\secspace
A Boolean network is a directed acyclic graph (DAG) that maps an $n$-dimensional Boolean vector to an $m$-dimensional one, such that $\{0,1\}^n \rightarrow \{0,1\}^m$. The input and output vectors are known as primary inputs (PIs) and primary outputs (POs). And-Inverter Graphs (AIGs), containing only two-input AND gates and inverters, are commonly used to represent Boolean networks due to their simplicity and scalability \cite{aig}.

Logic synthesis transformations can be classified into two categories: local operators, which are fast but achieve limited optimization, and global operators, which are slower but capable of more significant PPA improvements. Local operators only consider a node or small cut, resulting in a runtime linear to the cut size. In contrast, global resynthesis operators consider the entire graph when identifying optimization opportunities. Although global operators can substantially improve PPA, their prohibitive runtime makes them impractical for large industrial circuits.

We focus on the refactor operator, as its an essential component in industrial optimization flows and there has been little effort in literature to improve. While, refactor provides a lower area reduction relative to other operators, its usage is critical since its transformations unlock further optimizations in the AIG \cite{misch_2007}. Nevertheless, its running time is significant. For example, in a popular logic synthesis flow such as Resyn2, the refactor step typically accounts for $20-40\%$ of the total execution time, despite being invoked only twice, whereas the balance and rewrite steps are invoked three and four times, respectively. Amaru et al. proposed a Boolean filtering for these operators to reduce the number of gates process without affecting the quality of result (QoR) \cite{amaru_2018}. However in their results they do not report the runtime for the operators, so we cannot evaluate the effectiveness of the method.

\begin{enumerate}
    \item \textbf{Refactor} \cite{misch_2007} iterates over all the nodes and forms a large cut for each one of them. Each cut is transformed into an SOP and factored into a more compact form, while maintaining the function of the root node. If there are more removed nodes than nodes added, the new resynthesized subgraph gets committed. As refactor has seen minimal follow-on work since its proposal, research opportunities likely remain.

    \item \textbf{Rewrite} \cite{misch_2006} greedily iterates over all nodes, forming all 4-input cuts. It substitutes each cut with a precomputed minimal subgraph selected from 222 NPN classes. The subgraph with the maximum non-negative gain, i.e. removed nodes minus added nodes, is committed. Recent extensions consider 5-input subgraphs \cite{5input}, apply rewrite in a sliding window \cite{riener_2022}, and improvements in the efficiency by utilizing an engineered dynamic-scoring function \cite{Effisyn}.
    
    \item \textbf{Resubstitution} \cite{misch_2007} attempts to express the function of a node with other nodes present in the network called divisors. A transformation is accepted if the new implementation is better than the current node implementation using the immediate fanins. Resubstitution can be generalized to $k$-resubstitution. Let $k$ be the exact number of nodes added and $l$ the nodes within the MFFC of the node. If $k$-resubstitution is possible and $l>k$, the change is committed as the total number of nodes in the graph decreased. Since the graph is an AIG in our setting, the newly added nodes are two-input AND gates with potential inverters in the inputs or outputs. 
\end{enumerate}

\secspace
\section{Methodology} \label{sec:method}
\secspace
Section \ref{sec:glo} presents a general perspective on logic optimization operators and in more detail the refactor operator, motivating the need for our pruned approach, ELF. Section \ref{sec:pruning} delves into the technical details of the proposed pruning method, ELF, as well as optimizations to enhance performance. Finally, Section \ref{sec:feat_sel} enumerates the features uses for the classifier along with their rationale.

\subsection{Logic Optimization Operators \& Motivation}\label{sec:glo}
There is a range of logic optimization operators such as rewrite, refactor, and resubstitution which differ in their local transformations but, their algorithmic frameworks share key similarities. Each operator iterates through nodes in the same sequence, applying a distinct function to each node. Algorithm \ref{alg:gen_bool_resyn} illustrates the general pipeline of logic optimization.

\begin{algorithm}[!b]
    \caption{General form of Logic Optimization} \label{alg:gen_bool_resyn}
    \begin{algorithmic}[1]
            \FOR{each node in AIG}
            \STATE Apply Boolean\_Optimization(node, AIG)
            \STATE $Gain = \text{nodes removed} - \text{nodes added}$
            \IF{$Gain > 0$}
            \STATE Commit change in AIG
            \ENDIF
            \ENDFOR
    \end{algorithmic}
\end{algorithm}

In Algorithm \ref{alg:gen_bool_resyn}, the Boolean\_Optimization function resynthesizes the cut with fewer nodes, but commits the change only if the gain, defined as removed nodes minus added nodes, is positive. After cut optimization, functionality must be preserved by adding the necessary nodes. However, in most iterations, the cut fails resynthesis, wasting computation time. Specifically in refactor on the arithmetic circuits in EPFL arithmetic benchmark, at most $7.34\%$ of cuts are refactored. Similar trends hold for other operators, which can be even more time-intensive. Therefore, we propose ELF, an efficient learned classifier that given structural cut information, predicts whether a cut can get refactored. A key component behind ELF are the hand-crafted lightweight features that capture salient structural information to create compact cut representations.


\begin{figure}[!tb]
\centering
\includegraphics[width=\columnwidth]{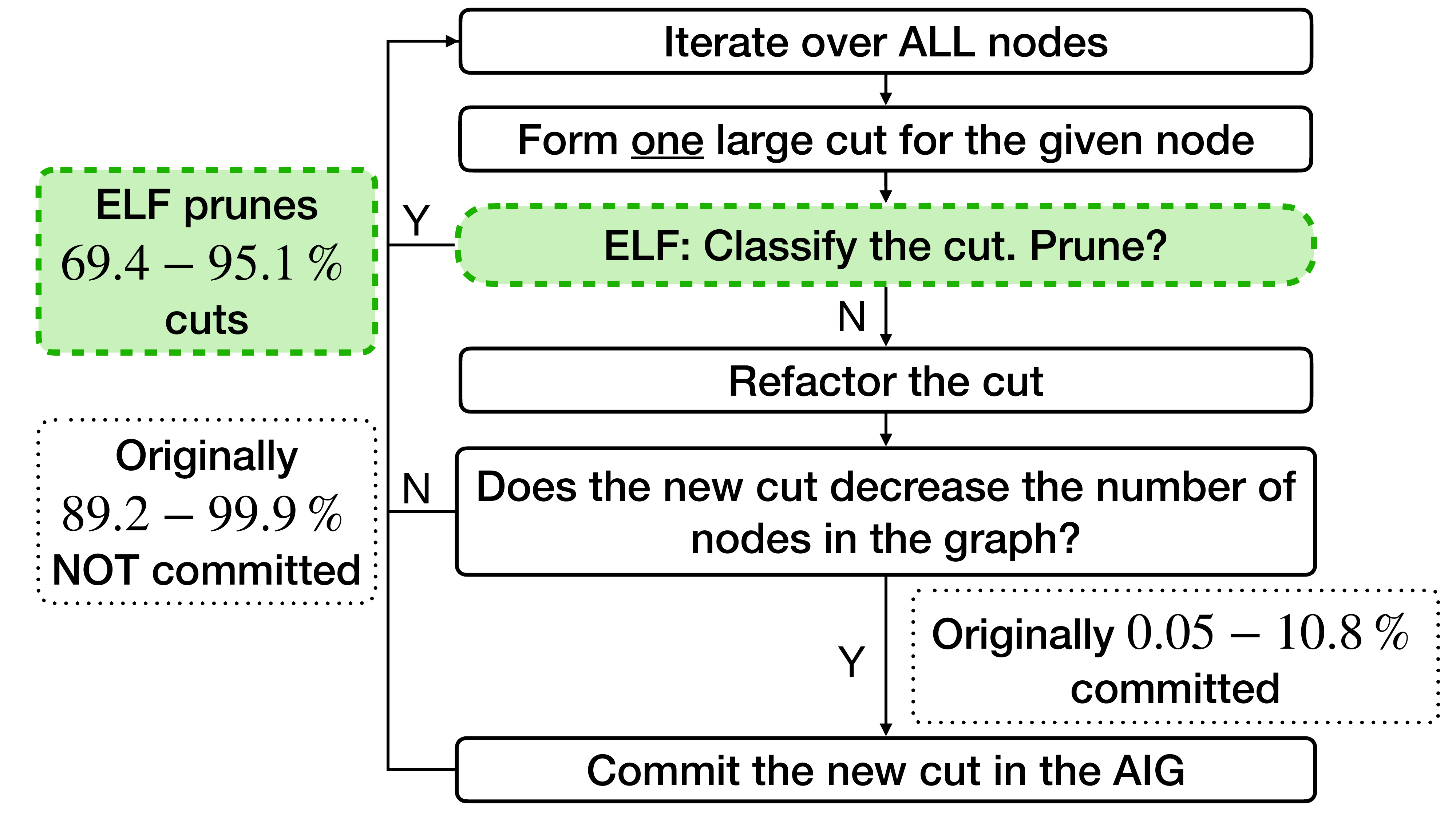}
\caption{Flow of ELF and the original refactor demonstrating the redundancy.} 
\label{fig:flow}
\end{figure}

\subsecspace
\subsection{ELF: Pruning Cuts}\label{sec:pruning}
\subsecspace

As shown in the Algorithm \ref{alg:gen_bool_resyn}, refactor iterates over all nodes in the And-Inverter Graph (AIG), forms a cut for each node and resynthesizes it. However, approximately $89.2 - 99.9\%$ of the formed cuts may fail to get optimized, wasting computational effort. We address this by classifying each node's benefit from resynthesis prior to function application. Specifically, we predict whether a given node is unlikely to be optimized, pruning it from the search space before proceeding. Only beneficial nodes undergo attempted resynthesis. Figure \ref{fig:flow} demonstrates the flow of the original ABC implementation of refactor with its shortcomings and the green cells highlight how ELF changes the flow and prevents unnecessary computation. ELF can be very effective as it can omit to refactor $69.4-95.1\%$ of the nodes.

At the core of our method, ELF, is an efficient and compact classifier. While refactor is time-consuming on large industrial circuits, it operates efficiently on a per-cut basis. This characteristic necessitates rapid inference to effectively reduce overall execution time. The stringent time constraints inherently disqualify most advanced model architectures, such as graph neural networks, for this application. To illustrate, the inference time per cut using a small Graph Convolutional Network (GCN) is approximately 30 times that of resynthesizing the cut itself.

After extensive experimentation with various model architectures, including XGBoost trees and diverse ensemble models, we found that a feedforward neural network offers the optimal balance between execution speed and generalization to unseen circuits, while maintaining high recall and accuracy. The computational complexity of a single layer in a feedforward neural network is $\Theta(m*h + k*h)$, where $m$ represents the number of inputs, $k$ the number of outputs and $h$ the number of hidden units.

Achieving high recall is crucial for our classifier, as our goal is to reduce runtime without compromising area optimization. Recall quantifies the percentage of nodes correctly classified as $1$. A recall of $100\%$ ensures no degradation in the quality of logic optimization. This emphasis on recall aligns with our objective of enhancing computational efficiency while preserving the effectiveness of the refactor operator in logic synthesis workflows. Equally crucial is high accuracy, which directly correlates with runtime reduction. As we accurately prune more nodes, we expect a proportional decrease in runtime. The synergy between high recall and accuracy thus forms the cornerstone of our approach, enabling efficient circuit optimization without sacrificing quality.

 To further decrease the inference time we use some engineering tricks such as kernel fusion and graph compilation for the model. Additionally, before node iteration, we batch all the cut data into one tensor so that we can maximize vectorization and minimize the process of packing/unpacking the data into tensors. Therefore, the complexity of classifying $N$ nodes with a model that has complexity $\Theta(M)$ is $\Theta(NM)$. Despite concerns about diminished accuracy, experiments validate the viability of the batch classification approach. An intuitive explanation for this phenomenon is that the cuts at the start, are more likely to have available optimizations, which might be lost while refactoring the AIG. Although, we might unnecessarily resynthesize cuts that no longer can be optimized, this will only impose a degradation in the runtime, but not the area. The detailed implementation of ELF is outlined in Algorithm \ref{alg:prune_bool_resyn}. 

\begin{algorithm}[!bt]
    \caption{ELF: Efficient Logic Refactoring} \label{alg:prune_bool_resyn}
    \begin{algorithmic}[1]
        \STATE Collect features for each node's cut
        \STATE Classify each node
            \FOR{each node in $AIG$}
                \IF{the node is classified to fail}
                    \STATE Continue
                \ENDIF
                \STATE  Refactor(node, AIG)
                \STATE $Gain = \text{nodes removed} - \text{nodes added}$
                \IF{$Gain > 0$}
                    \STATE Commit change in AIG
                \ENDIF
            \ENDFOR
    \end{algorithmic}
\end{algorithm}

As seen in Algorithm \ref{alg:gen_bool_resyn} the logic optimization methods share an overall similar structure. Thus, it is possible to extend ELF to work for other operators other than refactor. The ultimate goal would be to have a unified  logic optimization operator where the classifier conducts multiclass prediction to select the Boolean resynthesis function for each node, including skipping non-beneficial nodes. Thus, streamlining the entire process of logic synthesis. 

\subsecspace
\subsection{Feature Selection}\label{sec:feat_sel}
\subsecspace
The features must add minimal runtime overhead while encapsulating salient information about the number of nodes to be removed and the logic sharing. Logic sharing refers to nodes that connect to nodes outside the cut. Meaning that their deletion is non-trivial as they affect the functionality of other parts of the circuit. A straightforward approach to represent the cut is to use the cut's root truth table. Although the truth table captures the functionality of the cut, it fails to encapsulate logic sharing. Additionally, large truth tables scale exponentially with the number of leaves, potentially incurring substantial computational costs. Logic optimization operators resynthesize first the cut as an isolated graph then, consider the impact of reinserting the optimized cut. If the nodes that need to be added to maintain the logic sharing exceed the number of nodes removed, the cut does not get committed. Thus, structural cut information can potentially capture both node deletion opportunities and logic sharing information. 

\begin{figure}[!tb]
\centering
\includegraphics[width=\columnwidth]{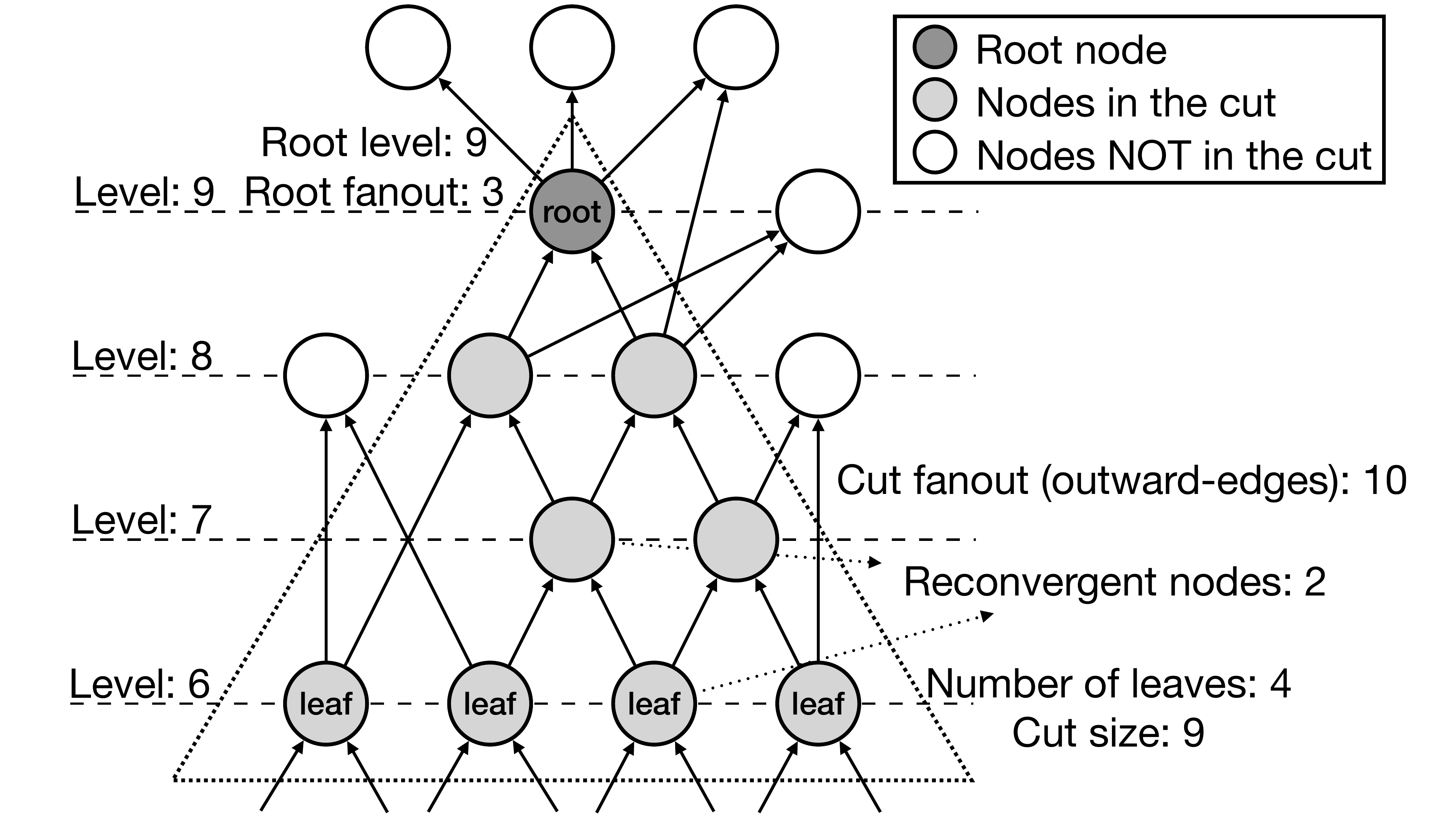}
\caption{Example of the features used to represent a cut in an AIG} 
\label{fig:aig}
\end{figure}

We represent cuts by 6-dimensional feature vectors, comprising: \textit{root fanout}, \textit{root level}, \textit{total cut fanout}, \textit{cut size}, \textit{number of reconvergent nodes}, and \textit{number of leaf nodes}. Figure \ref{fig:aig}, contains an example of a cut with its features. In refactor, given a node it forms a reconvergence driven cut, where the given node is the cut's root node. The \textit{root level} refers to the level of the given node within the AIG. The fanout refers to the outward going edges of a node. So, in Figure \ref{fig:aig} the cut's \textit{root level} is $9$ and has $3$ outgoing edges meaning the \textit{root fanout} is $3$. The triangle in the figure represents the cut and \textit{cut size} refers to the number of nodes within it, which is $9$ in this case. Similarly to the \textit{root fanout}, the \textit{total cut fanout} refers to the cut's total number of outgoing edges, which is $10$. For a node to be a leaf there must be at least one path from a primary input (PI) that goes through the leaf and reaches the root of the cut without going through any other leaf. In Fig. \ref{fig:aig}, there are 4 \textit{leaf nodes}. Finally, if there are two paths starting from a node $v_s$ and ending to the node $v_e$ they are denoted as reconvergent paths and $v_s$ is a reconvergent node. Note that we only count the nodes that are part of reconvergent paths locally within the cut, for simplicity. In Fig. \ref{fig:aig} there are only 2 locally \textit{reconvergent nodes} marked with the arrows. 

The rationale behind features related to the fanout are that they capture logic sharing as they indicate that they connect with nodes outside the cut. The bigger the fanout, the harder it is to replace the nodes within the cut. The \textit{cut size} indicates the number of nodes to potentially prune, but its role can be to regularize other features. For example, the neural network might learn to divide \textit{total cut fanout} by the \textit{cut size}. The leaves cannot be simplified so a higher \textit{number of leaves} might be less likely to be successfully optimized. The \textit{root’s level} implies the importance of a node. The higher level of a node, the more important it might be. Finally, in Theorem 1 in a recent paper by Riener et al. \cite{riener_2022}, they proved that if a certain type of optimization exists, then reconvergent nodes should exist. Thus, we include the number of \textit{reconvergent nodes} as a feature.

These lightweight features are ideal for our problem as they are cheap to compute. We achieve efficient collection by accumulating the features during cut construction. Our flexible approach can run on CPUs or parallelized to run on GPUs.

\secspace
\section{Experimental Results} \label{sec:exp results}
\secspace
In section \ref{sec:experimental settings}, we introduce the implementation details and information about the datasets. Section \ref{sec:comparisons} compares the efficiency and effectiveness of the ELF over the baseline method, the original state-of-the-art refactor implementation of ABC, followed by Section \ref{sec:model_quality}, which presents the experiments on the classifier's quality metrics. Finally, Section \ref{sec:features} contains explainability experiments along with an analysis of the features used and their impact on the model's output.

\subsecspace
\subsection{Experimental Settings}\label{sec:experimental settings}
\subsecspace
We evaluate on the EPFL arithmetic benchmarks \cite{amaru_2015} and industrial designs. While the results are consistent for all the circuits, we show the results for the 6 largest EPFL circuits, and train the model on all the AIGs in the suite except the one we test on. Similarly for the industrial circuits, we train on all the other designs except the one we test on for 10 designs and report the average results. Note, we keep the two datasets separate for reproducibility. Table \ref{tab:epfl_datasets} and Table \ref{tab:industrial_datasets} contains statistics for the AIGs in the EPFL benchmarks and industrial designs, on which we report detailed results. The \textit{Refactored} column the percentage represents the ratio of nodes that get resynthesized. It is evident that the datasets are highly imbalanced, since percentage of nodes that do get refactored are mostly around $~1\%$. Although these circuits vary in size, demonstrating the scalability of our method, they are smaller than real industrial circuits, which often contain tens of millions of nodes and exhibit similar trends. Only three AIGs contain a relatively higher number of nodes that can get refactored, \textit{sqrt} in the EPFL benchmarks and two industrial designs \textit{design} $5$ and $10$, which contain $7.34\%$, $10.8\%$, and $4.53\%$ nodes that can be resynthesized, respectively. This imbalance makes the learning significantly more difficult. Additionally, our classifier needs to be recall driven. Meaning that we do not want to misclassify cuts, since it will impact the quality of the operator. High accuracy is relatively easy to achieve, in the current setting, simply by classifying every cut as a $0$. The difficulty in our task is to find a good balance between recall and accuracy. Thus, minimizing running time, with minimal impact on the number of AND gates.

\begin{table}[!b]
    \figspace
    \caption{Arithmetic EPFL circuits statistics}
    \figtitlespace
    \figspace
    \begin{center}
    \begin{tabular}{c|c|c|c|c|c}
    \hline
    \textbf{Design} & \textbf{And} & \textbf{Level} & \textbf{PIs} & \textbf{POs} & \textbf{Refactored}  \\
    \hline
    div & 57247 & 4372 & 128 & 128 & 285 $(0.50\,\%)$\\
    \hline
    hyp & 214335 & 24801 & 256 & 128 & 1992 $(0.93\,\%)$\\
    \hline
    log2 & 32060 & 444 & 32 & 32 & 530 $(1.65\,\%)$\\
    \hline
    multiplier & 27062 & 274 & 128 & 128 & 247 $(0.91\,\%)$\\
    \hline
    sqrt & 24618 & 5058 & 128 & 64 & 1806 $(7.34\,\%)$\\
    \hline
    square & 18484 & 250 & 64 & 128 & 177 $(0.96\,\%)$\\
    \hline
    \end{tabular}
    \label{tab:epfl_datasets}
    \end{center}
    \figspace
\end{table}

\begin{table}[!b]
    \figspace
    \caption{Industrial circuits statistics}
    \figtitlespace
    \figspace
    \begin{center}
    \begin{tabular}{c|c|c|c|c|c}
    \hline
    \textbf{Design} & \textbf{And} & \textbf{Level} & \textbf{PIs} & \textbf{POs} & \textbf{Refactored}  \\
    \hline
    design 1 & 384971 & 65 & 13135 & 13127 & 1142 $(0.30\,\%)$\\
    \hline
    design 2 & 267358 & 49 & 27800 & 20603 & 1184 $(0.44\,\%)$\\
    \hline
    design 3 & 628777 & 36 & 35552 & 34480 & 1569 $(0.25\,\%)$\\
    \hline
    design 4 & 159763 & 44 & 35784 & 34712 & 1273 $(0.80\,\%)$\\
    \hline
    design 5 & 428904 & 51 & 52344 & 51283 & 46376 $(10.8\,\%)$\\
    \hline
    design 6 & 507027 & 35 & 26292 & 25220 & 603 $(0.12\,\%)$\\
    \hline
    design 7 & 305218 & 72 & 20228 & 19148 & 839 $(0.28\,\%)$\\
    \hline
    design 8 & 77130 & 40 & 18357 & 18325 & 42 $(0.05\,\%)$\\
    \hline
    design 9 & 190600 & 71 & 26168 & 26139 & 807 $(0.42\,\%)$\\
    \hline
    design 10 & 423661 & 40 & 42257 & 33849 & 19180 $(4.53\,\%)$\\
    \hline
    \end{tabular}
    \label{tab:industrial_datasets}
    \end{center}
    \figspace
\end{table}

To show that the models can generalize to unseen data, we train the classifier to all the arithmetic datasets other than the one we are testing on. Similarly for the industrial design circuits, none of the circuits that get tested are part of training. Additionally, each dataset is standardized individually with mean variance normalization. This way we can make the model infer to sizes that it might have not seen so far. The features we use to train the classifier with, can be calculated with simple accumulators while forming the cut with almost no additional runtime cost. This is necessary as refactor is already an efficient algorithm, leaving little running time available for inference. The features we use are \textit{root fanout}, \textit{root level}, \textit{cut total fanout}, \textit{cut size}, \textit{number of reconvergent nodes}, and \textit{number of leaf nodes}.

The experiments were run on an Intel(R) Xeon(R) Platinum 8180M CPU @ 2.50GHz with 1 thread and 1TB of RAM. The models were trained using pytorch and the model was run directly in C++ within ABC \cite{abc_2010} and executed with the ONNX Runtime framework. To efficiently standardize and execute the model we merged a Mean Variance Normalization node directly with the model and run all the cut data into one batch. While ML libraries are quite optimized, they have quite a bit of overhead when it comes to wrapping and unwrapping the data into tensors. Taking advantage of vectorization and limiting the overhead of the tensor objects, inference is very fast.

The model has 4 fully-connected layers with shape $6\rightarrow 12 \rightarrow 12 \rightarrow  6 \rightarrow 1$ and a total of 325 parameters. Although more sophisticated architectures could potentially achieve better results, they are often prohibited by the tight time-budget. For instance, convolutional layers could be employed to enable the model to learn feature extraction autonomously, but this would require a GPU to fully leverage their capabilities. We initialize the parameters of each layer with Xavier initialization and the biases set to 0. For training we used a batch size of 64, 30 epochs up with early stopping and patience of 10 epochs. The optimizer was Adam with learning rate of 0.1 and a learning rate scheduler of cosine annealing with warm restarts. While we tried a range of losses such as focal loss, class balanced loss and class balanced focal loss, binary cross entropy worked the best. Additionally, to augment the data, we used MixUp \cite{mixup} to improve generalization and overall model metrics. We tested more data augmentation techniques such as Synthetic Minority Over-sampling \cite{smote} and One Sided Selection \cite{oss}, but a weighted random sampler proved better. Furthermore, we tested dropout layers to potentially improve generalization, but it proved ineffective.

\subsecspace
\subsection{Comparison with Baseline} \label{sec:comparisons}
\subsecspace
To demonstrate the effectiveness of ELF we compare it against the original implementation of refactor within ABC \cite{abc_2010}. Additionally, we call the \textit{rf -l} version of the operator. The quality metrics we report include the number of AND gates, levels, and their relative differences, the runtimes and the speedup of the proposed method. Let the relative difference be $\delta$ for a metric $m_{\text{ELF}}$ for the proposed method, ELF, and $m_{\text{ABC}}$ for the original implementation of refactor in ABC. Then, we measure the differences as \( \delta=\frac{m_{\text{ELF}}-m_{\text{ABC}}}{m_{\text{ABC}}}\times 100\%.\) From an industrial perspective, the area degradation should not exceed $0.5\%$, while achieving a $1.25\times$ speedup.

\begin{table*}[!ht]
    \figspace
    \caption{Refactor's performance in original ABC vs ELF. Experiments on arithmetic circuits in the EPFL benchmark.}
    \figtitlespace
    \figspace
    \begin{center}
    \resizebox{0.95\linewidth}{!}{
    \begin{tabular}{c|c||c|c|c||c|c|c||c|c|c}
    \hline
    \multicolumn{2}{c||}{ } & \multicolumn{3}{c||}{Original ABC} & \multicolumn{3}{c||}{ELF (this work)} & \multicolumn{3}{c}{Difference} \\
    \hline
    \textbf{Design} & \textbf{Nodes} & \textbf{Runtime} ($s$)& \textbf{And} & \textbf{Level} & \textbf{Runtime} ($s$)& \textbf{And} & \textbf{Level} & \textbf{Speedup}& \textbf{And}& \textbf{Level} \\
    \hline
    div & 57247 & 1.00 & 56745 & 4372 & 0.21 & 56897 & 4372 &  4.76\,$\times$ & $+0.27\,\%$ & $0\,\%$ \\
    \hline
    hyp & 214335 & 6.82 & 212341 & 24801 & 0.93 & 212341 & 24801 & 7.33\,$\times$ & $\ \ \;0.00\,\%$ & $0\,\%$ \\
    \hline
    log2 & 32060 & 0.71 & 31517 & 445 & 0.13 & 31564 & 445 &  5.46\,$\times$ & $+0.15\,\%$ & $0\,\%$ \\
    \hline
    multiplier & 27062 & 0.77 & 26814 & 274 & 0.10 & 26815 & 274 & 7.69\,$\times$ & $\ \ \;0.00\,\%$ & $0\,\%$ \\
    \hline
    sqrt & 24618 & 0.25 & 22811 & 5932 & 0.10 & 22872 & 5931 &  2.50\,$\times$ & $+0.27\,\%$ & $0\,\%$ \\
    \hline
    square & 18484 & 0.52 & 18302 & 250 & 0.13 & 18314 & 250 & 4.00\,$\times$ & $+0.07\,\%$ & $0\,\%$ \\
    \hline
    \end{tabular}
    }
    \label{tab:quality_epfl}
    \end{center}
    \figspace
\end{table*}

\begin{table*}[htpb]
    \figspace
    \caption{Refactor's performance in original ABC vs applying ELF twice. Experiments on arithmetic circuits in the EPFL benchmark.}
    \figtitlespace
    \figspace
    \begin{center}
    \resizebox{0.95\linewidth}{!}{
    \begin{tabular}{c|c||c|c|c||c|c|c||c|c|c}
    \hline
    \multicolumn{2}{c||}{ } & \multicolumn{3}{c||}{Original ABC} & \multicolumn{3}{c||}{ELF $\times$ 2 \text{ (this work applied twice)}} & \multicolumn{3}{c}{Difference} \\
    \hline
    \textbf{Design} & \textbf{Nodes} & \textbf{Runtime} ($s$)& \textbf{And} & \textbf{Level} & \textbf{Runtime} ($s$)& \textbf{And} & \textbf{Level} & \textbf{Speedup}& \textbf{And}& \textbf{Level} \\
    \hline
    div & 57247 & 1.00 & 56745 & 4372 & 0.43 & 56664 & 4372 & 2.32\,$\times$ & $-0.14\,\%$ & $0\,\%$  \\
    \hline
    hyp & 214335 & 6.82 & 212341 & 24801 & 2.02 & 212296 & 24801 & 3.38\,$\times$ & $-0.02\,\%$ & $0\,\%$  \\
    \hline
    log2 & 32060 & 0.71 & 31517 & 445 & 0.53 & 31564 & 445 & 1.34\,$\times$ & $+0.15\,\%$ & $0\,\%$  \\
    \hline
    multiplier & 27062 & 0.77 & 26814 & 274 & 0.35 & 26814 & 274 & 2.20\,$\times$ &  $\ \ \;0.00\,\%$ & $0\,\%$ \\
    \hline
    sqrt & 24618 & 0.25 & 22811 & 5932 & 0.17 & 22872 & 5931 & 1.47\,$\times$ & $+0.27\,\%$ & $0\,\%$ \\
    \hline
    square & 18484 & 0.52 & 18302 & 250 & 0.27 & 18307 & 250 & 1.93\,$\times$ & $+0.03\,\%$ & $0\,\%$  \\
    \hline
    \end{tabular}
    }
    \label{tab:quality_x2_epfl}
    \end{center}
    \figspace
\end{table*}

\begin{table*}[!ht]
    \figspace
    \caption{Refactor's performance in original ABC vs ELF. Experiments on industrial circuits.}
    \figtitlespace
    \figspace
    \begin{center}
    \resizebox{0.95\linewidth}{!}{
    \begin{tabular}{c|c||c|c|c||c|c|c||c|c|c}
    \hline
    \multicolumn{2}{c||}{ } & \multicolumn{3}{c||}{Original ABC} & \multicolumn{3}{c||}{ELF (this work)} & \multicolumn{3}{c}{Difference} \\
    \hline
    \textbf{Design} & \textbf{Nodes} & \textbf{Runtime} ($s$)& \textbf{And} & \textbf{Level} & \textbf{Runtime} ($s$)& \textbf{And} & \textbf{Level} & \textbf{Speedup}& \textbf{And}& \textbf{Level} \\
    \hline
    design 1 & 384971 & 4.10 & 382941 & 54 & 1.32 & 383021 & 54 & 3.10\,$\times$ & $+0.02\,\%$ & $\ \ \,0.00\,\%$ \\
    \hline
    design 2 & 267358 & 2.95 & 265387 & 48 & 0.85 & 265582 & 49 &  3.47\,$\times$ & $+0.07\,\%$ & $+2.08\,\%$ \\
    \hline
    design 3 & 628777 & 6.35 & 627059 & 36 & 1.91 & 627059 & 36 &  3.32\,$\times$ & $\ \ \,0.00\,\%$ & $\ \ \,0.00\,\%$ \\
    \hline
    design 4 & 159763 & 1.63 & 157097 & 44 & 0.38 & 157228 & 44 &  4.29\,$\times$ & $+0.08\,\%$ & $\ \ \,0.00\,\%$ \\
    \hline
    design 5 & 428904 & 5.03 & 294992 & 46 & 2.17 & 295168 & 48 &  2.32\,$\times$ & $+0.06\,\%$ & $+4.35\,\%$ \\
    \hline
    design 6 & 507027 & 5.75 & 506416 & 36 & 2.32 & 506420 & 36 &  2.48\,$\times$ & $\ \ \,0.00\,\%$ & $\ \ \,0.00\,\%$ \\
    \hline
    design 7 & 305218 & 3.31 & 303609 & 64 & 1.48 & 303680 & 64 &  2.24\,$\times$ & $+0.02\,\%$ & $\ \ \,0.00\,\%$ \\
    \hline
    design 8 & 77130 & 0.82 & 77082 & 40 & 0.33 & 77082 & 40 &  2.48\,$\times$ & $\ \ \,0.00\,\%$ & $\ \ \,0.00\,\%$ \\
    \hline
    design 9 & 190600 & 1.75 & 189471 & 71 & 0.77 & 189520 & 68 &  2.27\,$\times$ & $+0.03\,\%$ & $-4.23\,\%$ \\
    \hline
    design 10& 423661 & 4.93 & 381423 & 39 & 2.45 & 381433 & 39 &  2.01\,$\times$ & $\ \ \,0.00\,\%$ & $\ \ \,0.00\,\%$ \\
    \hline
    \end{tabular}
    }
    \label{tab:industrial}
    \end{center}
    \figspace
\end{table*}

\begin{table}[!ht]
    \figspace
    \caption{Refactor's performance in original ABC vs ELF. Experiments on large synthetic circuits.}
    \figtitlespace
    \figspace
    \begin{center}
    \resizebox{0.99\linewidth}{!}{
    \begin{tabular}{c|c||c|c|c}
    \hline
    \textbf{Design} & \textbf{Nodes} & \textbf{ABC Runtime} ($s$) & \textbf{ELF Speedup}& \textbf{And Diff.} \\
    \hline
    sixteen & 16216836 &  2243.63 & 2.97\,$\times$ & $+0.07\,\%$ \\
    \hline
    twenty & 20732893 & 3138.46 & 2.87\,$\times$ & $+0.06\,\%$ \\
    \hline
    twentythree & 23339737 & 3914.77 & 2.85\,$\times$ & $+0.06\,\%$ \\
    \hline
    \end{tabular}
    }
    \label{tab:synthetic}
    \end{center}
    \figspace
\end{table}

Table \ref{tab:quality_epfl} presents the results of applying the original implementation of refactor operator within ABC vs this work, ELF, on the arithmetic circuits in the EPFL benchmark. The experiments show a significant runtime speedup varying from $2.50\times$ up to $7.69\times$. Simultaneously, there is minimal increase in the area. In the worst case, the number of nodes can increase by $0.27\%$ for \textit{div}, while the area difference is $0\%$ for other circuits. The level is the same among all circuits. While, the runtime of the refactor operator is generally small for the circuits in the EPFL benchmark, industrial designs might contain several million nodes making the original implementation expensive i.e. several hours. We later present experiments on large synthetic AIGs.

The significant runtime speedup enables us to apply ELF twice without exceeding the execution of the original implementation of refactor in ABC. Thus, Table \ref{tab:quality_x2_epfl} contains the results of chaining ELF twice against the baseline. In the worst case, ELF maintained a $1.34\times$ speedup in runtime, but it failed to improve the quality further. Overall, a subsequent application of the proposed method can improve the results except in two circuits, log2 and sqrt. It is important to note that three of the circuits, multiplier, sqrt, and square, do not have any potential improvements in general when applying the original refactor operator twice. In contrast, for \textit{div} and \textit{hyp} by applying ELF twice we can still have a significant speedup in the running time and reduce the area of the AIG. Note that, these two graphs are the two largest arithmetic circuits in the EPFL benchmark.

Table \ref{tab:industrial} shows that ELF is equally effective in 10 industrial designs. These circuits are larger than the ones within the EPFL benchmark and vary from 77k-629k nodes. The trend from the results on the circuits in the EPFL benchmark continues to industrial design that are larger. The runtime speedup varies from $2.01\times$ to $4.29\times$, while the added number of nodes is minuscule. In the worst case the number of nodes increased by $0.08\%$, but was also as low as $0\%$. No information about these 10 circuits was part of the training data, which demonstrates that the classifier succeeded in generalizing. Thus, demonstrating the effectiveness of ELF and potential applicability to industrial solutions.

Finally, Table \ref{tab:synthetic} presents the condensed results of applying the original implementation of refactor operator within ABC vs this work, ELF, on synthetic circuits in the EPFL benchmark. Originally, the runtime of refactor can be longer than 1 hour for circuits such as the one with 23 million nodes, but ELF can theoretically speedup the operator by $2.90\times$ on average with little impact on the And nodes difference. To effectively observe these speedups in practice for large circuits, it is necessary to invest engineering efforts into improving the implementation of replacing the refactored cut back into the AIG.

\subsecspace
\subsection{Model Quality Experiments} \label{sec:model_quality}
\subsecspace
To effectively prune cuts our classifier needs to be recall-driven first and also achieve a high enough accuracy. Recall and accuracy are calculated as \[\text{Recall} = \frac{\text{TP}}{\text{TP} + \text{FN}}, \ \ \text{Accuracy} = \frac{\text{TP} + \text{TN}}{\text{TP} + \text{FP} + \text{FN} + \text{TN}}.\] Recall directly affects the area, as it measures the proportion of nodes that were correctly classified as $1$. Instead, accuracy is related to the overall runtime. The more accurate the model the more nodes the algorithm can skip. Thus, reducing the running time.

Table \ref{tab:epfl_model_stats} contains the model quality metrics of ELF on the arithmetic circuits in the EPFL benchmark. The classifier achieves a high recall and accuracy, while it was never trained on the test circuit at all. Recall is consistently above $93\%$ and up to $100\%$ for some circuits, but it is only $76\%$ for \textit{div}. Finally, accuracy is overall high, varying from $77\%$ up to $96\%$, even though the recall is also high. Therefore, this justifies the runtime reduction, while maintaining a low impact on the area, varying from $0\%$ to $+0.27\%$.

\begin{table}[!ht]
    \figspace
    \caption{Model quality metrics of ELF on the arithmetic circuits in the EPFL benchmark.}
    \figtitlespace
    \figspace
    \begin{center}
    \resizebox{\columnwidth}{!}{%
    \begin{tabular}{c||c|c||c|c|c|c}
    \hline
    \textbf{Design} & \textbf{Recall} & \textbf{Accuracy} & \textbf{TP} & \textbf{TN} & \textbf{FP} & \textbf{FN} \\
    \hline
    div & $\ \;76\,\%$ & $84\,\%$ & $217$ & $48045$ & $8905$ & $68$ \\
    \hline
    hyp & $100\,\%$ & $77\,\%$ & $1992$ & $162061$ & $50282$ & $0$ \\
    \hline
    log2 & $\ \;93\,\%$ & $90\,\%$ & $494$ & $28295$ & $3235$ & $36$ \\
    \hline
    multiplier & $100\,\%$ & $96\,\%$ & $246$ & $25740$ & $1075$ & $1$ \\
    \hline
    sqrt & $\ \;97\,\%$ & $92\,\%$  & $1745$ & $20951$ & $1860$ & $61$ \\
    \hline
    square & $\ \;94\,\%$ & $84\,\%$  & $167$ & $15422$ & $2885$ & $10$ \\
    \hline
    \end{tabular}
    }
    \label{tab:epfl_model_stats}
    \end{center}
    \figspace
\end{table}

Table \ref{tab:industrial_model_stats} contains quality metrics of the ELF classifier. The results are consistent with the ones on the arithmetic circuits. Recall is high varying from $81\%$ to $100\%$ and so is accuracy $81-93\%$. We used the same classifier to test the quality on over 170 industrial designs. Recall is on average $91.8\%$ and accuracy $81.0\%$.

\begin{table}[!ht]
    \figspace
    \caption{Model metrics of ELF on the industrial circuits}
    \figtitlespace
    \figspace
    \begin{center}
    \resizebox{\columnwidth}{!}{%
    \begin{tabular}{c||c|c||c|c|c|c}
    \hline
    \textbf{Design} & \textbf{Recall} & \textbf{Accuracy} & \textbf{TP} & \textbf{TN} & \textbf{FP} & \textbf{FN} \\
    \hline
    design 1 & $\ \,94\,\%$ & $92\,\%$ & $1069$ & $351154$ & $32477$ & $73$ \\
    \hline
    design 2 & $\ \,81\,\%$ & $85\,\%$ & $958$ & $226380$ & $39711$ & $226$ \\
    \hline
    design 3 & $100\,\%$ & $93\,\%$ & $1564$ & $582528$ & $44604$ & $5$ \\
    \hline
    design 4 & $\ \,89\,\%$ & $93\,\%$ & $1132$ & $147701$ & $10780$ & $141$ \\
    \hline
    design 5 & $100\,\%$ & $81\,\%$ & $46272$ & $310728$ & $70552$ & $104$ \\
    \hline
    design 6 & $100\,\%$ & $87\,\%$ & $600$ & $439699$ & $66644$ & $3$ \\
    \hline
    design 7 & $\ \,91\,\%$ & $79\,\%$ & $762$ & $240122$ & $63985$ & $77$ \\
    \hline
    design 8 & $100\,\%$ & $79\,\%$ & $42$ & $60857$ & $16211$ & $0$ \\
    \hline
    design 9 & $\ \,94\,\%$ & $85\,\%$ & $755$ & $160845$ & $28924$ & $52$ \\
    \hline
    design 10 & $100\,\%$ & $74\,\%$ & $19144$ & $293832$ & $110537$ & $36$ \\
    \hline
    \end{tabular}
    }
    \label{tab:industrial_model_stats}
    \end{center}
    \figspace
\end{table}

\subsecspace
\subsection{Features Study}\label{sec:features}
\begin{figure}[!ht]
    \figspace
    \includegraphics[width=\linewidth]{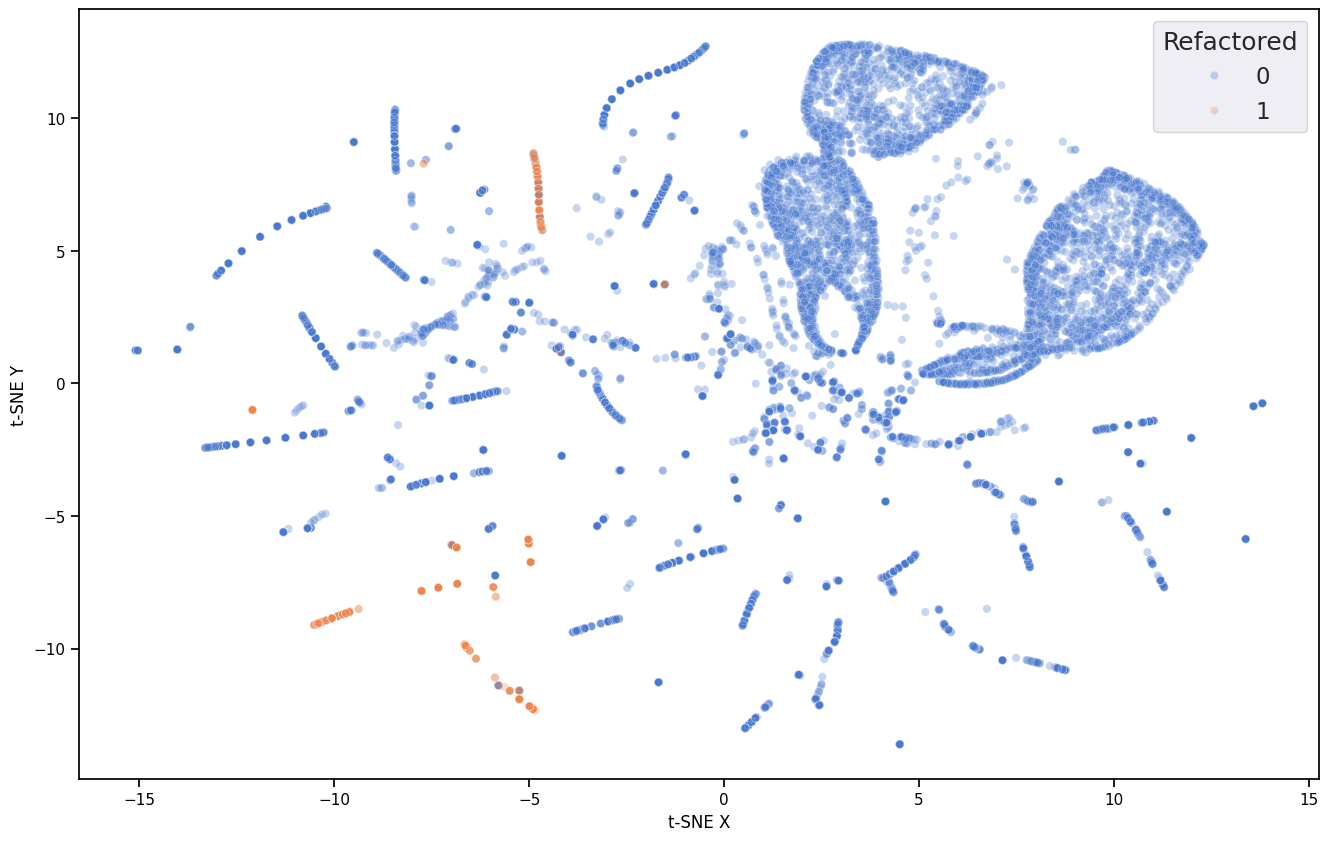}\\
    \figspace
\caption{t-SNE visualization of the feature space}
\label{fig:tsne}
\end{figure}

\subsecspace
t-SNE plots \cite{tsne} are a widely used technique for visualizing high-dimensional data in a two-dimensional space. Figure \ref{fig:tsne} presents such a visualization of our feature space, where orange points represent instances that underwent refactoring, and blue  points indicate those that did not. The plot reveals the presence of distinct clusters, albeit dispersed across the space. This cluster distribution can be attributed to three key factors: the highly imbalanced nature of our dataset, the relatively low dimensionality of our feature space, and the discrete characteristics of our features. Despite these challenges, the discernible separation between clusters suggests that effective classification is feasible.

\begin{figure}[!ht]
    \figspace
    \includegraphics[width=\linewidth]{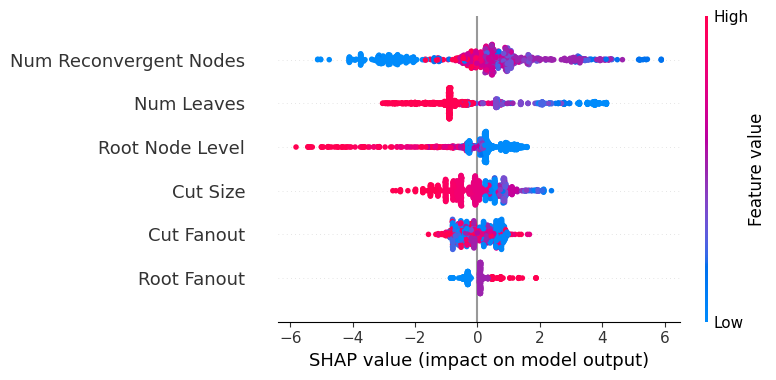}
    \figspace
    \figtitlespace
\caption{SHAP (SHapley Additive exPlanations) values for each feature in the model}
\label{fig:shap}
\end{figure}

Figure \ref{fig:shap} illustrates the SHAP \cite{shap} (SHapley Additive exPlanations) values for each feature in our model. SHAP values quantify the contribution of each input feature to the model's prediction, indicating both the magnitude and direction of influence. Positive SHAP values push the prediction towards 1 (indicating likely refactoring), while negative values push towards 0 (suggesting no refactoring). While SHAP values may not fully capture complex feature interactions, they provide valuable insights. For instance, the plot reveals that a low number of reconvergent nodes decreases the likelihood of refactoring, which aligns with the intuition that reconvergent nodes present optimization opportunities. Conversely, cuts with a high number of leaves, high root node level, or large cut size are less likely to be refactored. These features, while potentially not individually explanatory, can collectively convey important information about the cut. For example, a high leaf count in a moderately sized cut might indicate lower node redundancy, as leaf nodes are irreplaceable. The cut fanout shows ambiguous SHAP values, but the difference between the cut's fanout and the root's fanout represents the number of nodes without external edges, which are crucial for maintaining AIG functionality and thus cannot be replaced.

\secspace
\section{Conclusion} \label{sec:conclusion}
\secspace
In summary we propose ELF a pruning strategy for the refactor operator that can speedup resynthesis by $3.9\times$ on average with a minimal effect on the area. To achieve this we trained a compact model to classify, if the cuts will get successfully refactored. The features used to train the classifier are lightweight and ideal for time sensitive ML applications. An extensive experimental analysis on the EPFL benchmark suite and industrial circuits demonstrates the effectiveness in reducing the running time of the refactor operator. Given the similar structure of other logic synthesis operators such as rewrite and resubstituion, ELF can be adjusted to prune the unnecessary computation of those functions. A further extension of this work is to propose a unified logic optimization operation. Instead of having a sequence of operators, the classifier can choose the function to apply directly to the node or skip it. Such an operator could potentially significantly reduce the running time, achieve better PPA and eliminate the need to perform sequence tuning.

\newpage

\bibliographystyle{ieeetr}
\bibliography{bib.bib}

\end{document}